# Reliable Prediction Errors for Deep Neural Networks Using Test-Time Dropout

Isidro Cortés-Ciriano[1,*] and Andreas Bender[1]

[1]Centre for Molecular Informatics, Department of Chemistry, University of Cambridge, Lensfield Road, Cambridge, CB2 1EW, United Kingdom.

[*]Corresponding author: isidrolauscher@gmail.com

## Abstract

While the use of deep learning in drug discovery is gaining increasing attention, the lack of methods to computate reliable errors in prediction for Neural Networks prevents their application to guide decision making in domains where identifying unreliable predictions is essential, *e.g*. precision medicine. Here, we present a framework to compute reliable errors in prediction for Neural Networks using Test-Time Dropout and Conformal Prediction. Specifically, the algorithm consists of training a *single* Neural Network using dropout, and then *applying it N times* to both the validation and test sets, also employing dropout in this step. Therefore, for each instance in the validation and test sets an ensemble of predictions were generated. The residuals and absolute errors in prediction for the validation set were then used to compute prediction errors for test set instances using Conformal Prediction. We show using 24 bioactivity data sets from ChEMBL 23 that dropout Conformal Predictors are valid (*i.e.,* the fraction of instances whose true value lies within the predicted interval strongly correlates with the confidence level) and efficient, as the predicted confidence intervals span a narrower set of values than those computed with Conformal Predictors generated using Random Forest (RF) models. Lastly, we show in retrospective virtual screening experiments that dropout and RF-based Conformal Predictors lead to comparable retrieval rates of active compounds. Overall, we propose a computationally efficient framework (as only *N* extra forward passes are required in addition to training a single network) to harness Test-Time Dropout and the Conformal Prediction framework, and to thereby generate reliable prediction errors for deep Neural Networks.

## Introduction

The application of deep learning in preclinical and clinical applications is gaining increasing attention[1–4]. While alternative algorithms used in computer-aided drug design are still widely used, such as Random Forest (RF) or Support Vector Machines, the versatility of deep learning in terms of the diversity of architectures or feature representations that can be used as input (*e.g.,* from compound fingerprints to images of compounds[5,6], cells[7], or whole-slide pathology images[8,9]) renders it a suitable modelling strategy in diverse drug discovery applications[10–13]. However, the difficulty in interpreting *how deep learning models work* in most cases and the lack of methods to generate well-calibrated errors in prediction makes it challenging to apply Deep Neural Networks (DNN) to guide decision making in applications were knowing the reliability of individual predictions is essential *e.g.,* personalized medicine[14,15].

The development of techniques for the estimation of the reliability of individual predictions is a major research area in machine learning, and it is of particular importance in the area of drug discovery, where both the predicted property and the associated uncertainty of this prediction needs to be considered for decision making[16]. A plethora of algorithmic approaches have been proposed to this end, including metrics based on the distance of test instances to those in the training set[17,18], uncertainty estimation using Bayesian inference[19–24], and methods based on the bagged variance across base learners in model ensembles[25,26], among others[27–29]. In the context of deep learning, harnessing Test-Time Dropout predictions to model predictions uncertainty has gained increasing attention over the last years[23,30]. The main idea consists of training a Neural Network using dropout. Once trained, the network is applied on the test set to compute *N* forward passes using dropout as well. The variability across forward passes can be used to compute errors in prediction using Bayesian variational inference[23,30]. In addition to these Bayesian approaches[23,30], we recently showed that Snapshot Ensembles and Conformal Prediction can be integrated to generate reliable confidence intervals for deep learning architectures[31]. However, the development of algorithms to model predictions uncertainty using deep learning and Conformal Prediction in the context of regression still remains vastly unexplored[31].

The application of Conformal Prediction in drug discovery is becoming widely used[32–35] because: (i) the predicted confidence intervals are easy to interpret (*e.g.,* at a confidence level of 90%, at least 90% of the predicted confidence regions will contain the true value)[29]; (ii) the sound mathematical framework on which Conformal Prediction is based[36,37] guarantees that the confidence intervals are always *valid* (*i.e.,* the fraction of instances for which the true value lies within the predicted region will not be smaller than the chosen confidence level) provided that the exchangeability principle is fulfilled, which is generally assumed when modelling preclinical data[29,36,37], although this is not always the case[38]; (iii) each prediction consists of a confidence region in the case of regression, whose size depends on the user-defined Confidence Level (CL), and well-calibrated class probabilities in the case of classification[37]; (iv) Conformal Prediction can be used in combination with any machine learning algorithm and requires minimal computational cost beyond the training of the underlying algorithm[29]; (v) no parameterization is needed beyond the selection of a non-conformity function[39]. Hence, Conformal Prediction can be used in drug discovery to compute reliable errors in prediction at minimal computational cost.

In the current work, we explore the possibility of utilizing Test-Time Dropout ensembles to generate Conformal Predictors, which we term *Dropout Conformal Predictors*, with the aim of computing errors for individual predictions guaranteed to be valid and easy to interpret. The approach – in line with the Test-Time Dropout validations described above – consists of applying a deep Neural Network trained using dropout regularization to the validation and test molecules repeatedly in order to generate an ensemble. The residuals and variance across the ensemble for the validation set are then used to generate a list of nonconformity values, that in turn serve to generate confidence intervals for the test set molecules. Using 24 $IC_{50}$ data sets extracted from the ChEMBL database[40], we show that this strategy permits to generate valid and efficient Conformal Predictors. We show that the average size of the computed confidence intervals are comparable to those calculated using RF-based Conformal Predictors. In addition, we interrogate the practical usefulness of Dropout Conformal Predictors to guide the selection of active molecules in retrospective virtual screening experiments.

## Methods

### Data Sets

In this study we used 24 $IC_{50}$ data sets extracted from ChEMBL version 23[40], which have been previously modelled in several studies[31,38]. Briefly, we downloaded $IC_{50}$ data for 24 diverse protein targets from ChEMBL using the *chembl_webresource_client* python module[41–43]. Only $IC_{50}$ values for molecules that satisfied the following filtering criteria were considered: (i) activity relationship equal to '=', (ii) an activity unit equal to "nM", (iii) target type equal to "SINGLE PROTEIN", and (iv) organism equal to *Homo sapiens*. $IC_{50}$ values were modeled in a logarithmic scale ($pIC_{50} = -\log_{10} IC_{50}$). The average $pIC_{50}$ value was calculated when multiple $pIC_{50}$ values were available for the same compound. Further information about the data sets is given in Table 1 and [31,38]. The data sets are available in the Supporting Information of a previous study[38].

**Table 1. $IC_{50}$ data sets used in this study.**

| ChEMBL target preferred name | Target abbreviation | Uniprot ID | ChEMBL ID | Number of bioactivity data points |
|---|---|---|---|---|
| Tyrosine-protein kinase ABL | ABL1 | P00519 | CHEMBL1862 | 773 |
| Acetylcholinesterase | Acetylcholinesterase | P22303 | CHEMBL220 | 3,159 |
| Serine/threonine-protein kinase Aurora-A | Aurora-A | O14965 | CHEMBL4722 | 2,125 |
| Serine/threonine-protein kinase B-raf | B-raf | P15056 | CHEMBL5145 | 1,730 |
| Cannabinoid CB1 receptor | Cannabinoid | P21554 | CHEMBL218 | 1,116 |
| Carbonic anhydrase II | Carbonic | P00918 | CHEMBL205 | 603 |
| Caspase-3 | Caspase | P42574 | CHEMBL2334 | 1,606 |
| Thrombin | Coagulation | P00734 | CHEMBL204 | 1,700 |
| Cyclooxygenase-1 | COX-1 | P23219 | CHEMBL221 | 1,343 |
| Cyclooxygenase-2 | COX-2 | P35354 | CHEMBL230 | 2,855 |
| Dihydrofolate reductase | Dihydrofolate | P00374 | CHEMBL202 | 584 |
| Dopamine D2 receptor | Dopamine | P14416 | CHEMBL217 | 479 |
| Norepinephrine transporter | Ephrin | P23975 | CHEMBL222 | 1,740 |
| Epidermal growth factor receptor erbB1 | erbB1 | P00533 | CHEMBL203 | 4,868 |
| Estrogen receptor alpha | Estrogen | P03372 | CHEMBL206 | 1,705 |
| Glucocorticoid receptor | Glucocorticoid | P04150 | CHEMBL2034 | 1,447 |
| Glycogen synthase kinase-3 beta | Glycogen | P49841 | CHEMBL262 | 1,757 |
| HERG | HERG | Q12809 | CHEMBL240 | 5,207 |
| Tyrosine-protein kinase JAK2 | JAK2 | O60674 | CHEMBL2971 | 2,655 |
| Tyrosine-protein kinase LCK | LCK | P06239 | CHEMBL258 | 1,352 |
| Monoamine oxidase A | Monoamine | P21397 | CHEMBL1951 | 1,379 |
| Mu opioid receptor | Opioid | P35372 | CHEMBL233 | 840 |
| Vanilloid receptor | Vanilloid | Q8NER1 | CHEMBL4794 | 1,923 |

### Molecular Representation

All chemical structures were standardized to a common representation scheme using the python module *standardizer* (https://github.com/flatkinson/standardiser). Inorganic molecules were removed. The largest fragment was kept in order to filter out counterions, following standard procedures in the field[44,45]. Circular Morgan fingerprints[46] were computed using RDkit (release version 2013.03.02)[47]. The radius was set to 2 and the fingerprint length to 2,048.

## Machine Learning

### - Data Splitting

The data sets were randomly split into a training set (70% of the data), a validation set (15%), and a test set (15%). For each data set, the training set was used to train a given network, whereas the validation set served to monitor the performance of the network during the training phase. In case of RF models, both the training and validation sets were used for model training. The predictive power of the final RF and DNN model was evaluated on the test set. The above split (and associated model training and testing) was repeated 20 times with random data set assignments.

### - Deep Neural Networks (DNN)

DNNs were trained using the python library Pytorch[48]. We defined four hidden layers, composed of 1000, 1000, 100 and 10 nodes, respectively. The number of neurons in each layer was selected to be smaller than the input fingerprint size to reduce the chances of overfitting[49]. Rectified linear unit (ReLU) activation was used in all cases. The training data was processed in batches of size equal to 15% of the number of instances. We used Stochastic Gradient Descent with Nesterov momentum, which was set to 0.9 and kept constant during the training phase[50]. The networks were trained over 4,000 epochs, and early stopping was used in all cases, *i.e.,* the training of a given network was stopped if the validation loss did not decrease after 300 epochs. Cyclical learning rate annealing was implemented to train all networks[51,52]. Specifically, the learning rate was initially set to 0.005, and was decreased by 40% every 200 epochs. The learning rate was set back to the initial value (*i.e.,* 0.005) after 1,000 epochs and the annealing process repeated in the same way as done for the first 1,000 epochs. Setting back the learning rate to a high value permits to escape local minima by sampling larger regions of the loss landscape[51,52]. We used either 10%, 20% or 50% dropout in the four hidden layers both during training and at test time[2,53,54]. The RMSE on the validation set was used as the loss function during the training of all networks. DNNs that failed to converge to RMSE values on the validation set smaller than 1.2 $pIC_{50}$ units were discarded.

### - Random Forests (RF)

RF models were trained using the python library scikit-learn[55] as previously described[31,38]. Briefly, the default parameter values were used except for the number of trees, which was set to 100, as using a larger number of trees does not result in increased generalization capability and this value has been found to be a suitable choice in previous chemical structure-activity modelling studies[56,57].

## Conformal Prediction

- ***Dropout Conformal Prediction***

We describe below the steps followed to generate Conformal Predictors using Test-Time Dropout (Figure 1). We refer the reader to the work of Vovk *et al.*[29,36,37] for further theoretical details about the Conformal Prediction framework.

Firstly, a Neural Network was trained on the training and validation sets using dropout regularization[53] (step 1). Subsequently, it was applied to the validation set using Test-Time Dropout to generate 100 predictions for each instance. In each forward pass, the dropout probability of each neuron in each layer was independent of other passes (step 2). The predicted value for each instance in the validation set, $\hat{y}$, was then calculated as the average across the 100 forward passes, and the standard deviation across passes, $\sigma$, was used as a measurement of the prediction's uncertainty[58]. Next (step 3), the residuals and the standard deviation across forward passes were used to generate a list of non-conformity values for the validation set as follows:

$$\text{Equation 1:} \qquad \propto_i = \frac{|y_i - \hat{y}_i|}{e^{\sigma_i}}$$

where $y_i$ is the *i*$^{th}$ instance in the validation set, and $\hat{y}_i$ and $\sigma_i$ are the average and the standard deviation of the predicted activities for the *i*$^{th}$ instance across the 100 forward passes, respectively. The resulting list of non-conformity scores, $\alpha$, was sorted in increasing order, and the percentile corresponding to the Confidence Level considered was selected, *e.g.,* $\alpha_{80}$ for the 80$^{th}$ percentile.

The residuals were normalized using the natural exponential of the standard deviation across the 100 forward passes because comparative analysis of non-conformity metrics (as shown in Equation 1) revealed that the exponential scaling improves the efficiency of Conformal Predictors built on bioactivity data sets[39]. This scaling sets the upper value for the list of non-conformity values to be equal to the largest residual in the validation set, as the exponential converts low $\sigma_i$ values to values close to unity.

Finally, the standard deviation across forward passes was used to calculate confidence regions for the data points in the test set as follows (step 4 in Figure 1):

$$\text{Equation 2:} \quad Confidence\ region = \hat{y}_j \pm |y_j - \hat{y}_j| = \hat{y}_j \pm (e^{\sigma_j} * \alpha_{CL})$$

Where $y_j$ is the $j^{th}$ instance in the test set, $\hat{y}_j$ and $\sigma_j$ are the average and the standard deviation of the predicted activities for the $j^{th}$ instance across 100 forward passes, respectively. Throughout this work we considered a Confidence Level of 80% unless otherwise stated, as this confidence level in our experience represents a generally suitable trade-off between efficiency and validity[39].

- ***Random Forests***

Cross-Conformal Predictors generated using Random Forests were used as a baseline for comparison. RF-based Cross-Conformal Predictors were generated as previously reported[34,59]. Briefly, RF models were trained on the training data using 10-fold cross validation. The cross-validation residuals and the standard deviation across the Forest were used to calculate the list of non-conformity values for the training data[39,57], which were in turn used to compute confidence intervals for the test set instances.

## Results and Discussion

We firstly compared the performance on the test set of DNN, using dropout probabilities in all layers of either 0.1, 0.25 or 0.5, and RF models (Figure 2). Overall, the performance of RF and DNN (as well as performance for different dropout probabilities) was comparable, with RMSE values on the test set in the 0.55-0.80 $pIC_{50}$ units range, indicating that our choice of DNN architectures and model parameters were suitable to model these data sets. Moreover, the RMSE values on the test set are consistent with the expected errors in prediction for machine learning models trained on heterogeneous $IC_{50}$ data sets extracted from ChEMBL[60,61], and are in line with models reported in the literature for similar data sets[38]. Hence, the models obtained here are likely approaching the upper performance limit which can be obtained for the datasets used, which is also a likely factor behind the very similar performance obtained across methods.

In the Conformal Prediction literature, two main metrics are used to assess the practical usefulness of Conformal Predictors, namely *validity* and *efficiency*[29]. A Conformal Predictor is considered to be valid (or well calibrated) if the confidence level matches the fraction of data points in the test set whose true value lies within the predicted confidence region. For instance, at a confidence level of 80%, the confidence intervals computed using a valid Conformal Predictor would contain the true value in at least 80% of the cases. The Conformal Predictors we generated using Test-Time Dropout are well calibrated (*i.e.,* valid) for all dropout levels as well as for the RF-based models, as evidenced by the high correlation between the error rates and the confidence level ($R^2 > 0.99$, $P < 0.001$; Supplementary Figure 1).

Next, we evaluated the *efficiency* of the Conformal Predictors. The efficiency of a Conformal Predictor is determined by the size of the confidence regions. While a Conformal Predictor might be valid, it might not be useful in practice if the confidence regions span several $pIC_{50}$ units[39,62]. The *average* size of the confidence regions computed using RF-based and Dropout Conformal Predictors is comparable (Figure 3). However, the confidence intervals generated using Dropout Conformal Predictors span a smaller range of values. These results is line with previous results obtained for DNN-based Conformal Predictors generated using the Deep Confidence framework[31], which is based on Snapshot Ensembles[63]. Taken together, these results indicate that the ensembles generated using Test-Time Dropout can be harnessed to generate valid and efficient Conformal Predictors, with comparable validity and efficiency to RF-based ones, and fewer models associated with very large confidence intervals.

RF models are widely used in Conformal Predictors because the generation of an ensemble comes at no extra computational cost[29,64]. Moreover, the variance across base learners (*i.e.,* bagged variance across the Trees in the Random Forest) conveys predictive signal to quantify the uncertainty of individual predictions, as the average RMSE on the test set increases with the variance among predictions[25,26]. As reported previously[26], one should note that this numerical Pearson correlation between variance and prediction error is much weaker than the inflated correlation obtained by binning the test compounds on the basis of the predicted variance[25,26,65]. Most RF-based Conformal Predictors reported in the literature applied to regression tasks employ the standard deviation across base learners to scale the residuals and compute nonconformity scores[39,66] (Equation 1). In practice, this means that the average RMSE on the test set is correlated with the size of the confidence interval computed using Conformal Predictors because, *on average*, the larger the variance across the ensemble, the larger the predicted confidence region will be (Equation 2).

In the case of DNN, variational inference applied to Test-Time Dropout ensembles has been shown to generate well-calibrated confidence intervals, as their size strongly correlated with the average error in prediction[24]. In the current study we find that the bagged variance computed using Test-Time Dropout DNN ensembles spans a narrower range of values than the bagged variance calculated using RF (Figure 4 and Supplementary Figure 2), as evidenced by the larger spread of standard deviations between models (along the *x*-axis of Figure 4). A constant trend across data sets is that the higher the dropout probability, the higher the spread of standard deviations across models (Figure 4 and Supplementary Figure 2). Although Dropout Conformal Predictors generated using forward passes are valid, the smaller spread of variance values indicates that the size of the confidence intervals computed for test molecules with different absolute errors in prediction will be comparable. In other words, we observe a weaker correlation between the bagged variance and the absolute error in prediction for dropout DNN as compared to RF models. Hence, this indicates that the size of the predicted confidence intervals from dropout DNN models are *less* representative of the absolute error in prediction than those computed from RF models.

A major practical application of predicting uncertainty in a drug discovery setting is the prioritization of compounds for further experimental testing by selecting those with the lower bound of the predicted confidence region over a given bioactivity cut-off value[29]. For instance, if the goal was to find molecules with a potency better than 10nM ($pIC_{50}$>7), we would consider for further testing those molecules with a predicted confidence region that spans at least that value. Therefore, we next sought to investigate the practical usefulness of RF-based and Dropout Conformal Predictors to guide prospective virtual screening campaigns. To this aim, we defined as true positives (*i.e.,* the molecules we are interested in selecting) those molecules with the predicted lower bound of the confidence interval (*i.e.*, the lower end of the interval) above a given cut-off value (note that in real-world virtual screens also factors such as scaffold diversity etc.. are of relevance[67], which have not been considered in further detail in this analysis). Figure 5 shows the distribution of (i) uncertain predictions: molecules for which the confidence region spans the cut-off value, (ii) false positives: molecules whose true $pIC_{50}$ value is below the cut-off but the lower bound of the confidence region higher than the cut-off , (iii) true positives: molecules for which the lower bound of the confidence region and the true $pIC_{50}$ value are both over the cut-off (these are the molecules we would primarily prioritize for further experimental testing), and (iv) false negatives: molecules for which the upper bound of the confidence region is below the cut-off value but the true $pIC_{50}$ value is over the cut-off. We considered 5 integer-valued $pIC_{50}$ cut-off values in the analysis, ranging from 5 to 9.

Overall, both the total number and fraction of true positives that are discovered are comparable across algorithms, suggesting that Dropout Conformal Predictors permit the discovery of compounds with an activity value over a certain $pIC_{50}$ cut-off with comparable efficiency to RF ones, a trend that is observed across the $pIC_{50}$ cut-off values considered. A similar trend is observed for the number of false positives, false negatives and molecules whose predicted confidence region spans the cut-off value of interest. Together, these results indicate that a comparable number of molecules with a potency over a cut-off value of interest would have been discovered by application of RF and Dropout Conformal Predictors.

## Conclusions

In this work we have shown that Test-Time Dropout represents an approach to generate ensembles from the training of a single Neural Network which can be used to generate valid and efficient Conformal Predictors at minimal computational cost (*i.e.,* the cost of computing the extra forward passes). Comparable retrieval rates were obtained for RF-based and dropout Conformal Predictors in retrospective virtual screening experiments, where molecules with the predicted confidence intervals above a bioactivity cut-off value of interest were considered as active. Test-Time Dropout thus expands the set of algorithmic approaches available in preclinical drug discovery, which employs deep learning and is able to model the uncertainty of individual predictions.

## Figures

**Figure 1 Workflow for the generation of Dropout Conformal Predictors, illustrated by the training of a single DNN**. The first step (marked as 1 in the figure) consists of training a DNN using dropout. Subsequently, the network is applied to the validation set in 100 forward passes using random dropout (step 2). The residuals and variance across these 100 passes serve to compute a list of nonconformity scores (Equation 1). Next, the network is applied to the test set (step 3) 100 times using dropout as well. The resulting variance across forward passes is used to compute the point prediction and the confidence region (step 4 and Equation 2).

**Figure 2 Assessment of the predictive power on the test set.** Mean RMSE values (+/- standard deviation) for the test set molecules averaged across 20 runs are shown. The modelling approaches considered (RF and DNN trained using increasingly larger dropout probabilities) showed high predictive power, with mean RMSE values in agreement with the uncertainty of heterogeneous data sets extracted from ChEMBL. The comparable performance of the models suggests that they are likely approaching the upper performance limit which can be obtained for the datasets used.

**Figure 3 Efficiency analysis of the generated models**. Each box plot shows the distribution of the size of the confidence intervals generated for the test set molecules for a Confidence Level of 80% across 20 runs. Overall, the average interval size is comparable for Dropout and RF-based Conformal Predictors. However, the spread of the distributions for Dropout Conformal Predictors is smaller, indicating the absence of large confidence intervals.

**Figure 4 Relationship between the variance across forward passes and the absolute error in prediction.** Each panel represents the correlation between the standard deviation across forward passes (*x*-axis; standard deviation across the forest in the case of RF) against the absolute error in prediction (*y*-axis) for the test set instances**.** Results for the four data sets with the highest number of datapoints are shown: Acetylcholinesterase (A), erbB1 (B), HERG (C), and JAK2 (D). Overall, the base learners (*i.e.,* trees) in the case of RF models display a larger variation than DNN ensembles generated using Test-Time Dropout, as evidenced by the larger spread on the *x*-axis. Overall**,** the smaller spread of variance values in the case of DNN indicates that the size of the confidence intervals computed for molecules with different absolute errors in prediction will be more similar than those computed with RF-based Conformal Predictors.

**Figure 5 Comparison of retrieval rates for RF and Dropout Conformal Predictors.** Each column corresponds to a bioactivity cut-off value ($pIC_{50}$>5, 6, 7, 8, or 9). The bars represent the total number of test instances for which: (i) the confidence region spans the cut-off value (*i.e.,* uncertain predictions; shown in turquoise), (ii) the lower bound of the confidence region is higher than the cut-off value but the observed $pIC_{50}$ value is below the cut-off (*i.e.,* false positives, shown in yellow), (iii) the lower bound of the confidence region is higher than the cut-off value and the observed $pIC_{50}$ value is over the cut-off (*i.e.,* true positives; shown in purple), and (iv) the upper bound of the confidence region is below the cut-off value but the observed $pIC_{50}$ value is over the cut-off (*i.e.,* false negatives; shown in red). The numbers on top of the bars indicate the percentage of true positives (*i.e.,* molecules for which both the lower end of the confidence interval and the true $pIC_{50}$ value are above the cut-off). The average values across the 20 runs are shown. Overall, both the total number and fraction of true positives that are discovered are comparable across algorithms, suggesting that DNN Dropout Conformal Predictors permit the discovery of compounds with an activity value over a certain $pIC_{50}$ cut-off with comparable efficiency to RF ones while avoiding models with excessive prediction uncertainty.

## Supplementary Figures

**Supplementary Figure 1 Calibration plots.** Correlation between the estimated (1 - error rate) on the test set and the confidence level computed using (A) DNN dropout 0.1, (B) DNN dropout 0.25, (C) DNN dropout 0.5, and (D) RF models train on the data sets with the largest number of datapoints. Comparable results were obtained for the other data sets. Overall, Dropout Conformal Predictors were found to be valid and well calibrated, as evidenced by the high correlation between the confidence level and the percentage of predictions whose true value lies within the predicted confidence region (1- error rate; *y*-axis).

**Supplementary Figure 2 Analysis of the variance of DNN-based predictions across forward passes**. Distributions of the standard deviation across the ensemble for the test set instances are shown across the 20 runs. Overall, it can be seen that the variance across the ensemble is significantly larger for RF as compared to dropout-based DNN. This affects the spread of the distribution of intervals sizes, as shown in Figure 4.

## Author Contributions

I.C.-C. designed research, trained the models, and analyzed the results. I.C.-C. generated the figures. I.C.-C and A.B. wrote the paper.

## Acknowledgements

This project has received funding from the European Union's Framework Programme For Research and Innovation Horizon 2020 (2014-2020) under the Marie Curie Sklodowska-Curie Grant Agreement No. 703543 (I.C.C.).

## Conflicts of Interest

The authors declare no conflict of interests.

## References

(1) Chen, H.; Engkvist, O.; Wang, Y.; Olivecrona, M.; Blaschke, T. The Rise of Deep Learning in Drug Discovery. *Drug Discov. Today* **2018**, *23* (6), 1241–1250.

(2) Koutsoukas, A.; Monaghan, K. J.; Li, X.; Huan, J. Deep-Learning: Investigating Deep Neural Networks Hyper-Parameters and Comparison of Performance to Shallow Methods for Modeling Bioactivity Data. *J. Cheminform.* **2017**, *9* (1), 42.

(3) Unterthiner, T.; Mayr, A.; Unter Klambauer, G. ¨; Steijaert, M.; Wegner, J. K.; Johnson, J. &; Ceulemans, H.; Hochreiter, S. Deep Learning as an Opportunity in Virtual Screening.

(4) Lecun, Y.; Bengio, Y.; Hinton, G. Deep Learning. *Nature*. Nature Publishing Group May 28, 2015, pp 436–444.

(5) Goh, G. B.; Siegel, C.; Vishnu, A.; Hodas, N. O.; Baker, N. Chemception: A Deep Neural Network with Minimal Chemistry Knowledge Matches the Performance of Expert-Developed QSAR/QSPR Models. *2017, arXiv1706.06689 arXiv.org ePrint Arch. https//arxiv.org/abs/1706.06689 (accessed Jul 8, 2018).*

(6) Ciriano, I. C.; Bender, A. KekuleScope: Improved Prediction of Cancer Cell Line Sensitivity Using Convolutional Neural Networks Trained on Compound Images. **2018**.

(7) Hofmarcher, M.; Rumetshofer, E.; Clevert, D.-A.; Hochreiter, S.; Klambauer, G. Accurate Prediction of Biological Assays with High-Throughput Microscopy Images and Convolutional Networks. *J. Chem. Inf. Model.* **2019**, acs.jcim.8b00670.

(8) Yu, K.-H.; Berry, G. J.; Rubin, D. L.; Ré, C.; Altman, R. B.; Snyder, M. Association of Omics Features with Histopathology Patterns in Lung Adenocarcinoma. *Cell Syst.* **2017**, *5* (6), 620–627.e3.

(9) Coudray, N.; Ocampo, P. S.; Sakellaropoulos, T.; Narula, N.; Snuderl, M.; Fenyö, D.; Moreira, A. L.; Razavian, N.; Tsirigos, A. Classification and Mutation Prediction from Non–Small Cell Lung Cancer Histopathology Images Using Deep Learning. *Nat. Med.* **2018**, 1.

(10) Simm, J.; Klambauer, G.; Arany, A.; Steijaert, M.; Wegner, J. K.; Gustin, E.; Chupakhin, V.; Chong, Y. T.; Vialard, J.; Buijnsters, P.; Velter, I.; Vapirev, A.; Singh, S.; Carpenter, A. E.; Wuyts, R.; Hochreiter, S.; Moreau, Y.; Ceulemans, H. Repurposing High-Throughput Image Assays Enables Biological Activity Prediction for Drug Discovery. *Cell Chem. Biol.* **2018**, *25* (5), 611–618.e3.

(11) Duvenaud, D.; Maclaurin, D.; Aguilera-Iparraguirre, J.; Gómez-Bombarelli, R.; Hirzel, T.; Aspuru-Guzik, A.; Adams, R. P. *Convolutional Networks on Graphs for Learning Molecular Fingerprints*; 2015.

(12) Jiménez, J.; Škalič, M.; Martínez-Rosell, G.; De Fabritiis, G. $K_{\mathrm{DEEP}}$ : Protein–Ligand Absolute Binding Affinity Prediction via 3D-Convolutional Neural Networks. *J. Chem. Inf. Model.* **2018**, *58* (2), 287–296.

(13) Whitehead, T. M.; Irwin, B. W. J.; Hunt, P.; Segall, M. D.; Conduit, G. J. Imputation of Assay Bioactivity Data Using Deep Learning.

(14) Towards Trustable Machine Learning. *Nat. Biomed. Eng.* **2018**, *2* (10), 709–710.

(15) Yu, K.-H.; Beam, A. L.; Kohane, I. S. Artificial Intelligence in Healthcare. *Nat. Biomed. Eng.* **2018**, *2* (10), 719–731.

(16) Segall, M. D.; Champness, E. J. The Challenges of Making Decisions Using Uncertain Data. *J. Comput. Aided. Mol. Des.* **2015**, *29* (9), 809–816.

(17) Berenger, F.; Yamanishi, Y. A Distance-Based Boolean Applicability Domain for Classification of High Throughput Screening Data. *J. Chem. Inf. Model.* **2019**, *59* (1), 463–476.

(18) Liu, R.; Wallqvist, A. Molecular Similarity-Based Domain Applicability Metric Efficiently

Identifies Out-of-Domain Compounds. *J. Chem. Inf. Model.* **2019**, *59* (1), 181–189.
(19) Zhou, P.; Tian, F.; Chen, X.; Shang, Z. Modeling and Prediction of Binding Affinities between the Human Amphiphysin SH3 Domain and Its Peptide Ligands Using Genetic Algorithm-Gaussian Processes. *J. Pept. Sci.* **2008**, *90* (6), 792–802.
(20) Burden, F. R. Quantitative Structure-Activity Relationship Studies Using Gaussian Processes. *J. Chem. Inf. Comput. Sci.* **2001**, *41* (3), 830–835.
(21) Obrezanova, O.; Csányi, G.; Gola, J. M. R.; Segall, M. D. Gaussian Processes: A Method for Automatic QSAR Modeling of ADME Properties. *J. Chem. Inf. Model.* **2007**, *47* (5), 1847–1857.
(22) Cortes-Ciriano, I.; van Westen, G. J.; Lenselink, E. B.; Murrell, D. S.; Bender, A.; Malliavin, T. Proteochemometric Modeling in a Bayesian Framework. *J. Cheminf.* **2014**, *6* (1), 35.
(23) Zhang, Y.; Lee, A. A. *Bayesian Semi-Supervised Learning for Uncertainty-Calibrated Prediction of Molecular Properties and Active Learning*; 2019.
(24) Kendall, A.; Gal, Y. *What Uncertainties Do We Need in Bayesian Deep Learning for Computer Vision?*; 2017.
(25) Wood, D. J.; Carlsson, L.; Eklund, M.; Norinder, U.; Stålring, J. QSAR with Experimental and Predictive Distributions: An Information Theoretic Approach for Assessing Model Quality. *J. Comput. Aided Mol. Des.* **2013**, *27* (3), 203–219.
(26) Sheridan, R. P. Using Random Forest To Model the Domain Applicability of Another Random Forest Model. *J. Chem. Inf. Model.* **2013**, *53* (11), 2837–2850.
(27) Schroeter, T. S.; Schwaighofer, A.; Mika, S.; Ter Laak, A.; Suelzle, D.; Ganzer, U.; Heinrich, N.; Müller, K.-R. Estimating the Domain of Applicability for Machine Learning QSAR Models: A Study on Aqueous Solubility of Drug Discovery Molecules. *J. Comput. Mol. Des.* **2007**, *21* (9), 485–498.
(28) Sheridan, R. P. The Relative Importance of Domain Applicability Metrics for Estimating Prediction Errors in QSAR Varies with Training Set Diversity. *J. Chem. Inf. Model.* **2015**, *55* (6), 1098–1107.
(29) Norinder, U.; Carlsson, L.; Boyer, S.; Eklund, M.; Lundbeck, † H; Ottiliavej, A.; Denmark, V. Introducing Conformal Prediction in Predictive Modeling. A Transparent and Flexible Alternative To Applicability Domain Determination. *J. Chem. Inf. Model.* **2014**, *54* (6), 1596–1603.
(30) Gal, Y.; Ghahramani, Z. Dropout as a Bayesian Approximation: Representing Model Uncertainty in Deep Learning. *2015, arXiv1506.02142 arXiv.org ePrint Arch. https//arxiv.org/abs/1506.02142 (accessed Jul 10, 2018).*
(31) Cortés-Ciriano, I.; Bender, A. Deep Confidence: A Computationally Efficient Framework for Calculating Reliable Prediction Errors for Deep Neural Networks. *J. Chem. Inf. Model.* **2018**, acs.jcim.8b00542.
(32) Ahmed, L.; Georgiev, V.; Capuccini, M.; Toor, S.; Schaal, W.; Laure, E.; Spjuth, O. Efficient Iterative Virtual Screening with Apache Spark and Conformal Prediction. *J. Cheminform.* **2018**, *10* (1), 8.
(33) Svensson, F.; Norinder, U.; Bender, A. Improving Screening Efficiency through Iterative Screening Using Docking and Conformal Prediction. *J. Chem. Inf. Model.* **2017**, *57* (3), 439–444.
(34) Sun, J.; Carlsson, L.; Ahlberg, E.; Norinder, U.; Engkvist, O.; Chen, H. Applying Mondrian Cross-Conformal Prediction To Estimate Prediction Confidence on Large Imbalanced Bioactivity Data Sets. *J. Chem. Inf. Model.* **2017**, *57* (7), 1591–1598.
(35) Bosc, N.; Atkinson, F.; Felix, E.; Gaulton, A.; Hersey, A.; Leach, A. R. Large Scale Comparison of QSAR and Conformal Prediction Methods and Their Applications in Drug Discovery. *J. Cheminform.* **2019**, *11* (1), 4.
(36) Vovk, V.; Gammerman, A.; Shafer, G. *Algorithmic Learning in a Random World*; Springer,

2005.
(37) Shafer, G.; Vovk, V. A Tutorial on Conformal Prediction. *J. Mach. Learn. Res.* **2008**, *9*, 371–421.
(38) Cortes-Ciriano, I.; Firth, N. C.; Bender, A.; Watson, O. Discovering Highly Potent Molecules from an Initial Set of Inactives Using Iterative Screening. *J. Chem. Inf. Model.* **2018**, *58* (9), 2000–2014.
(39) Svensson, F.; Aniceto, N.; Norinder, U.; Cortes-Ciriano, I.; Spjuth, O.; Carlsson, L.; Bender, A. Conformal Regression for Quantitative Structure–Activity Relationship Modeling—Quantifying Prediction Uncertainty. *J. Chem. Inf. Model.* **2018**, *58* (5).
(40) Gaulton, A.; Hersey, A.; Nowotka, M.; Bento, A. P.; Chambers, J.; Mendez, D.; Mutowo, P.; Atkinson, F.; Bellis, L. J.; Cibrián-Uhalte, E.; Davies, M.; Dedman, N.; Karlsson, A.; Magariños, M. P.; Overington, J. P.; Papadatos, G.; Smit, I.; Leach, A. R. The ChEMBL Database in 2017. *Nucleic Acids Res.* **2017**, *45* (D1), D945–D954.
(41) Nowotka, M.; Papadatos, G.; Davies, M.; Dedman, N.; Hersey, A. Want Drugs? Use Python. *2016, arXiv1607.00378 arXiv.org ePrint Arch. https//arxiv.org/abs/1607.00378 (accessed Jul 10, 2018).*
(42) Davies, M.; Nowotka, M.; Papadatos, G.; Dedman, N.; Gaulton, A.; Atkinson, F.; Bellis, L.; Overington, J. P. ChEMBL Web Services: Streamlining Access to Drug Discovery Data and Utilities. *Nucleic Acids Res.* **2015**, *43* (W1), W612-20.
(43) Gaulton, A.; Bellis, L. J.; Bento, A. P.; Chambers, J.; Davies, M.; Hersey, A.; Light, Y.; McGlinchey, S.; Michalovich, D.; Al-Lazikani, B.; Overington, J. P. ChEMBL: A Large-Scale Bioactivity Database for Drug Discovery. *Nucleic Acids Res.* **2011**, *40* (D1), 1100–1107.
(44) Fourches, D.; Muratov, E.; Tropsha, A. Trust, but Verify: On the Importance of Chemical Structure Curation in Cheminformatics and QSAR Modeling Research. *J. Chem. Inf. Model.* **2010**, *50* (7), 1189–1204.
(45) O'Boyle, N. M.; Sayle, R. A. Comparing Structural Fingerprints Using a Literature-Based Similarity Benchmark. *J. Cheminform.* **2016**, *8* (1), 36.
(46) Rogers, D.; Hahn, M. Extended-Connectivity Fingerprints. *J. Chem. Inf. Model.* **2010**, *50* (5), 742–754.
(47) Landrum, G. RDKit: Open-Source Cheminformatics. *https//www.rdkit.org/ (accessed Jan 12, 2017)*.
(48) Paszke, A.; Chanan, G.; Lin, Z.; Gross, S.; Yang, E.; Antiga, L.; Devito, Z. Automatic Differentiation in PyTorch. In *Advances in Neural Information Processing Systems 30*; 2017; pp 1–4.
(49) Sheela, K. G.; Deepa, S. N. Review on Methods to Fix Number of Hidden Neurons in Neural Networks. *Math. Probl. Eng.* **2013**, *2013*, 1–11.
(50) Sutskever, I.; Martens, J.; Dahl, G.; Hinton, G. On the Importance of Initialization and Momentum in Deep Learning. In *Proceedings of the 30th International Conference on Machine Learning, PMLR 28*; 2013; pp 1139–1147.
(51) Smith, L. N. Cyclical Learning Rates for Training Neural Networks. In *2017 IEEE Winter Conference on Applications of Computer Vision (WACV)*; 2017; pp 464–472.
(52) Smith, L. N.; Topin, N. Exploring Loss Function Topology with Cyclical Learning Rates. *2017, arXiv1702.04283 arXiv.org ePrint Arch. https//arxiv.org/abs/1702.04283 (accessed Jul 8, 2018).*
(53) Srivastava, N.; Hinton, G.; Krizhevsky, A.; Salakhutdinov, R. Dropout: A Simple Way to Prevent Neural Networks from Overfitting. *J. Mach. Learn. Res.* **2014**, *15*, 1929–1958.
(54) Lenselink, E. B.; ten Dijke, N.; Bongers, B.; Papadatos, G.; van Vlijmen, H. W. T.; Kowalczyk, W.; IJzerman, A. P.; van Westen, G. J. P. Beyond the Hype: Deep Neural Networks Outperform Established Methods Using a ChEMBL Bioactivity Benchmark Set. *J. Cheminform.* **2017**, *9* (1), 45.

(55) Pedregosa, F.; Varoquaux, G.; Gramfort, A.; Michel, V.; Thirion, B.; Grisel, O.; Blondel, M.; Prettenhofer, P.; Weiss, R.; Dubourg, V.; Vanderplas, J.; Passos, A.; Cournapeau, D.; Brucher, M.; Perrot, M.; Duchesnay, E. Scikit-Learn: Machine Learning in Python. *J. Mach. Learn. Res.* **2011**, *12*, 2825–2830.
(56) Sheridan, R. P. Three Useful Dimensions for Domain Applicability in QSAR Models Using Random Forest. *J. Chem. Inf. Model.* **2012**, *52* (3), 814–823.
(57) Cortés-Ciriano, I.; van Westen, G. J. P.; Bouvier, G.; Nilges, M.; Overington, J. P.; Bender, A.; Malliavin, T. E. Improved Large-Scale Prediction of Growth Inhibition Patterns Using the NCI60 Cancer Cell Line Panel. *Bioinformatics* **2016**, *32* (1), 85–95.
(58) Beck, B.; Breindl, A.; Clark, T. QM/NN QSPR Models with Error Estimation: Vapor Pressure and LogP. *J. Chem. Inf. Comput. Sci.* **2000**, *40* (4), 1046–1051.
(59) Vovk, V. Cross-Conformal Predictors. *Ann. Math. Artif. Intell.* **2015**, *74* (1–2), 9–28.
(60) Kalliokoski, T.; Kramer, C.; Vulpetti, A.; Gedeck, P. Comparability of Mixed $IC_{50}$ Data - a Statistical Analysis. *PLoS One* **2013**, *8* (4), e61007.
(61) Cortés-Ciriano, I.; Bender, A. How Consistent Are Publicly Reported Cytotoxicity Data? Large-Scale Statistical Analysis of the Concordance of Public Independent Cytotoxicity Measurements. *ChemMedChem* **2015**, *11* (1), 57–71.
(62) Johansson, U.; Linusson, H.; Löfström, T.; Boström, H. Interpretable Regression Trees Using Conformal Prediction. *Expert Syst. Appl.* **2018**, *97*, 394–404.
(63) Cortes-Ciriano, I.; Bender, A. Deep Confidence: A Computationally Efficient Framework for Calculating Reliable Errors for Deep Neural Networks. *J. Chem. Inf. Model.* **2018**, *In press*, 1809.09060.
(64) Breiman, L. Random Forests. *Mach. Learn.* **2001**, *45* (1), 5–32.
(65) Wainer, H.; Gessaroli, M.; Verdi, M. Visual Revelations. *CHANCE* **2006**, *19* (1), 49–52.
(66) Cortés-Ciriano, I.; van Westen, G. J. P.; Bouvier, G.; Nilges, M.; Overington, J. P.; Bender, A.; Malliavin, T. E. Improved Large-Scale Prediction of Growth Inhibition Patterns on the NCI60 Cancer Cell-Line Panel. *Bioinformatics* **2016**, *32* (1), 85–95.
(67) Krier, M.; Guillaume Bret, A.; Rognan, D. Assessing the Scaffold Diversity of Screening Libraries. **2006**.

**Figure 1**

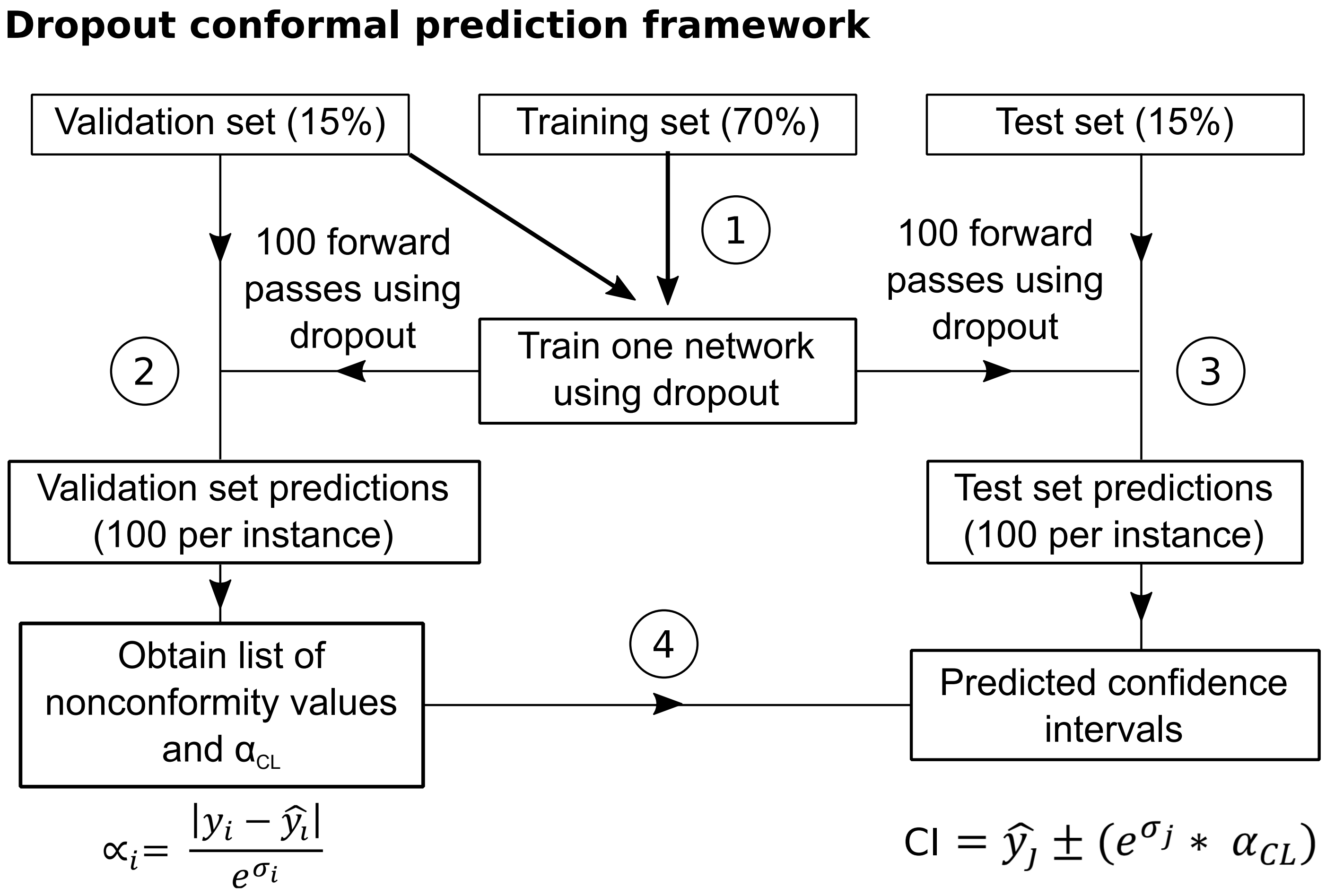

Dropout conformal prediction framework
Validation set (15%)
Training set (70%)
Test set (15%)
1
100 forward passes using dropout
100 forward passes using dropout
Train one network using dropout
2
3
Validation set predictions (100 per instance)
Test set predictions (100 per instance)
4
Obtain list of nonconformity values and $\alpha_{CL}$
Predicted confidence intervals
$\alpha_i = \frac{|y_i - \hat{y}_i|}{e^{\sigma_i}}$
$CI = \hat{y}_j \pm (e^{\sigma_j} * \alpha_{CL})$

**Figure 2**

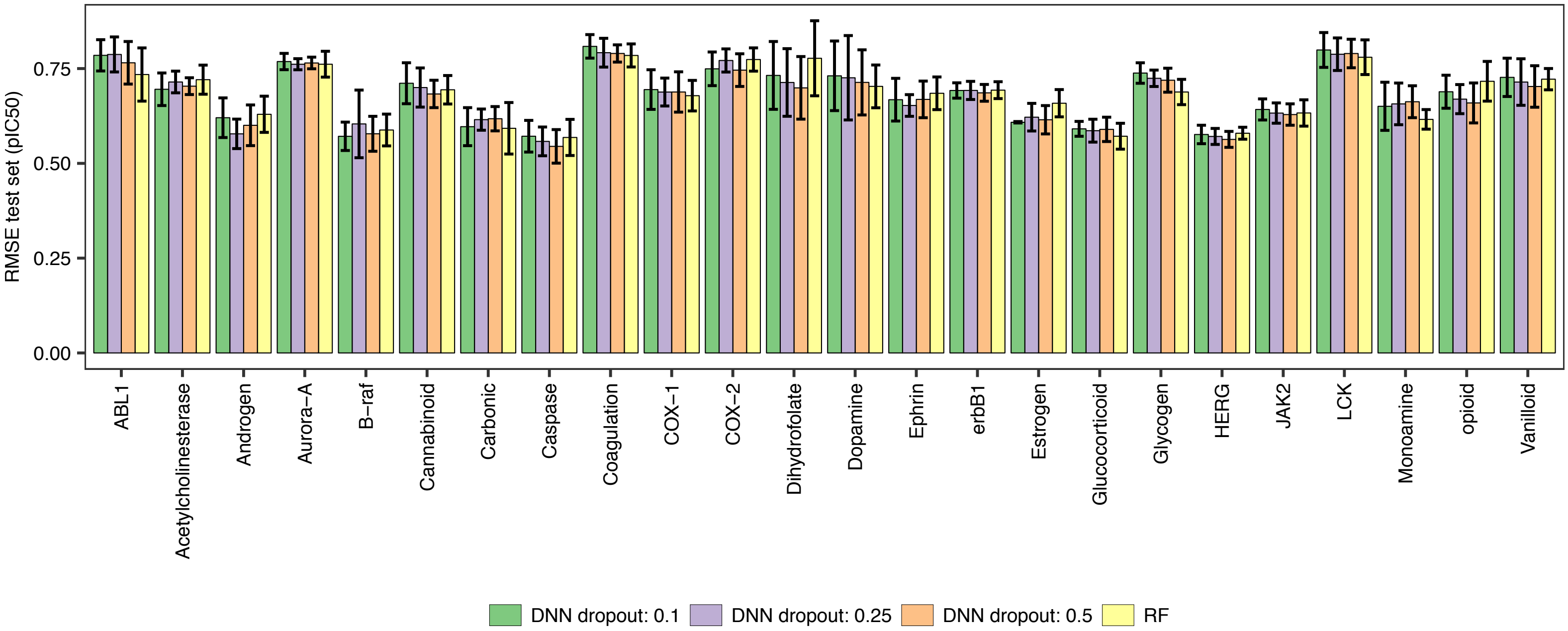

**Figure 3**

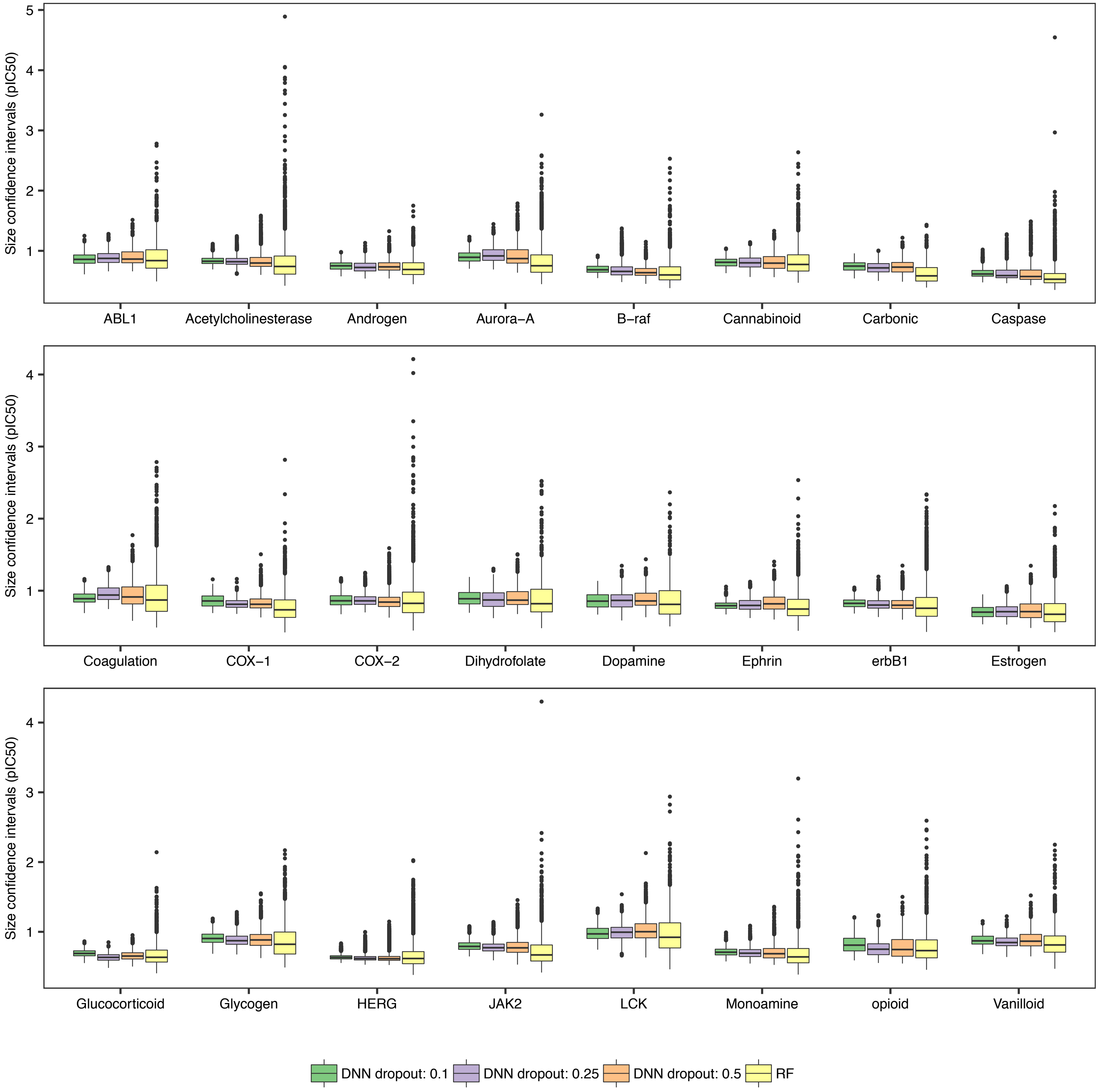

**Figure 4**

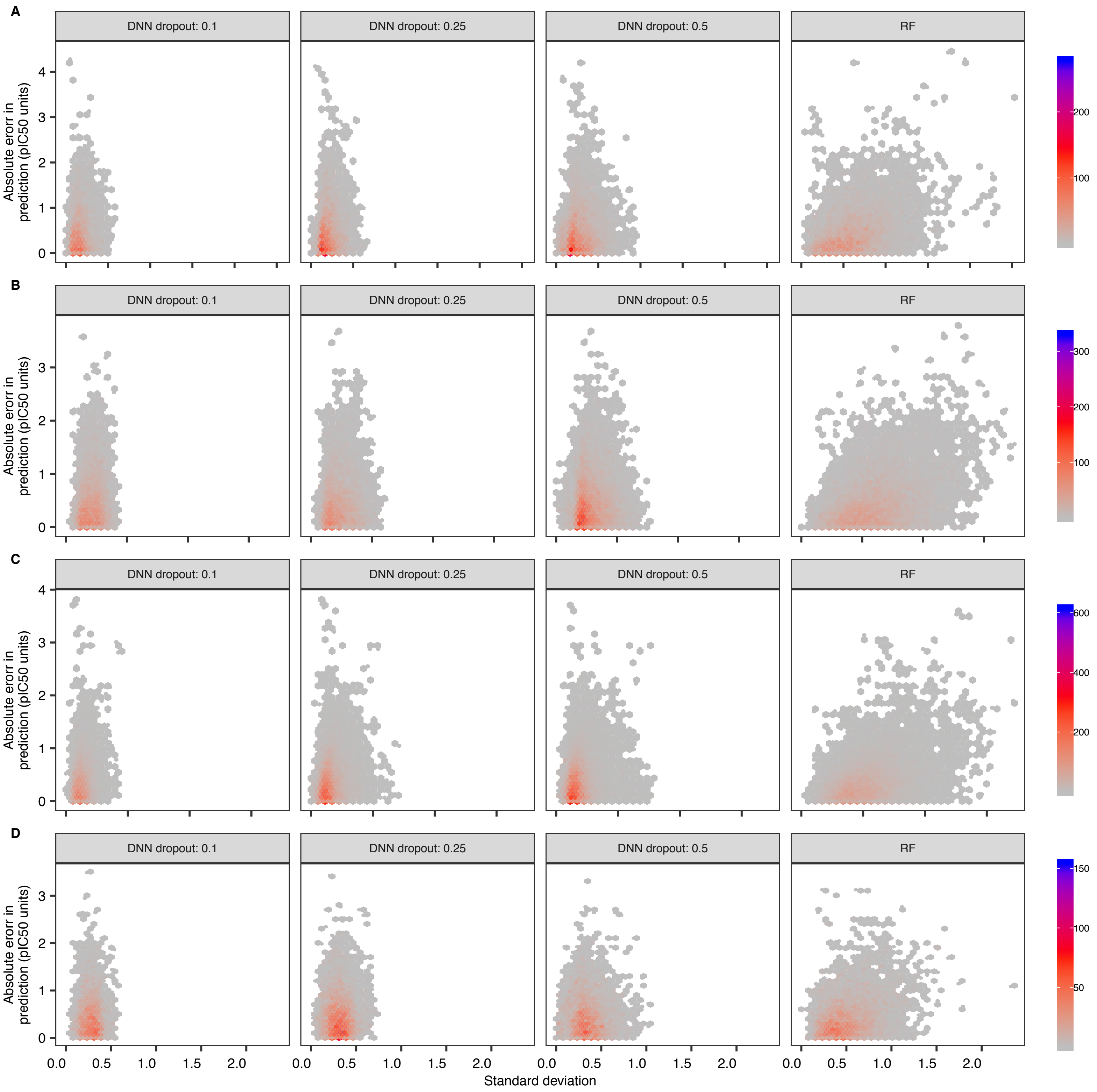

**Figure 5**

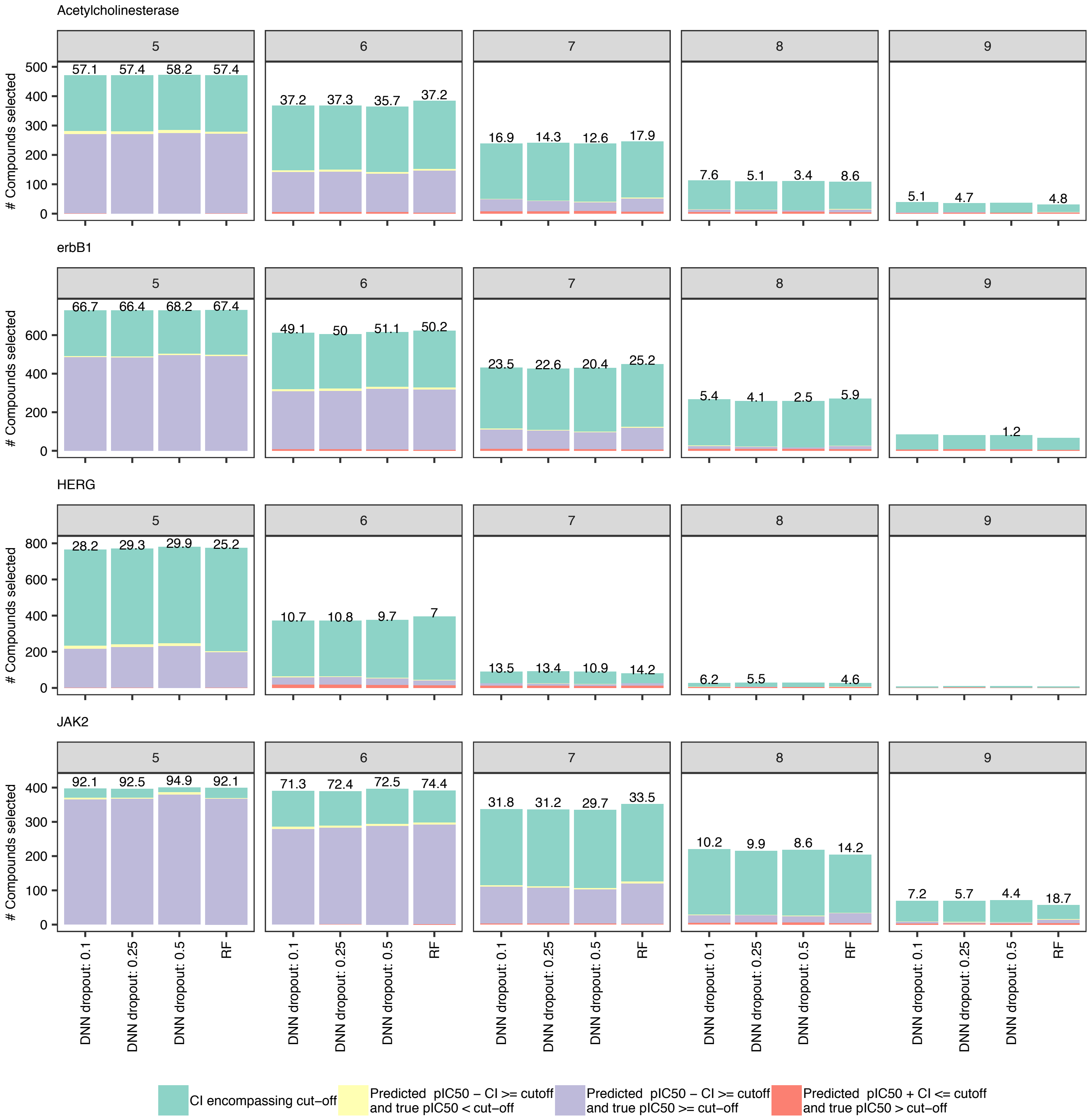

Acetylcholinesterase
5
6
7
8
9
# Compounds selected
57.1 57.4 58.2 57.4
37.2 37.3 35.7 37.2
16.9 14.3 12.6 17.9
7.6 5.1 3.4 8.6
5.1 4.7 4.8
erbB1
66.7 66.4 68.2 67.4
49.1 50 51.1 50.2
23.5 22.6 20.4 25.2
5.4 4.1 2.5 5.9
1.2
HERG
28.2 29.3 29.9 25.2
10.7 10.8 9.7 7
13.5 13.4 10.9 14.2
6.2 5.5 4.6
JAK2
92.1 92.5 94.9 92.1
71.3 72.4 72.5 74.4
31.8 31.2 29.7 33.5
10.2 9.9 8.6 14.2
7.2 5.7 4.4 18.7
DNN dropout: 0.1
DNN dropout: 0.25
DNN dropout: 0.5
RF
CI encompassing cut–off
Predicted pIC50 – CI >= cutoff and true pIC50 < cut–off
Predicted pIC50 – CI >= cutoff and true pIC50 >= cut–off
Predicted pIC50 + CI <= cutoff and true pIC50 > cut–off

**Supplementary Figure 1**

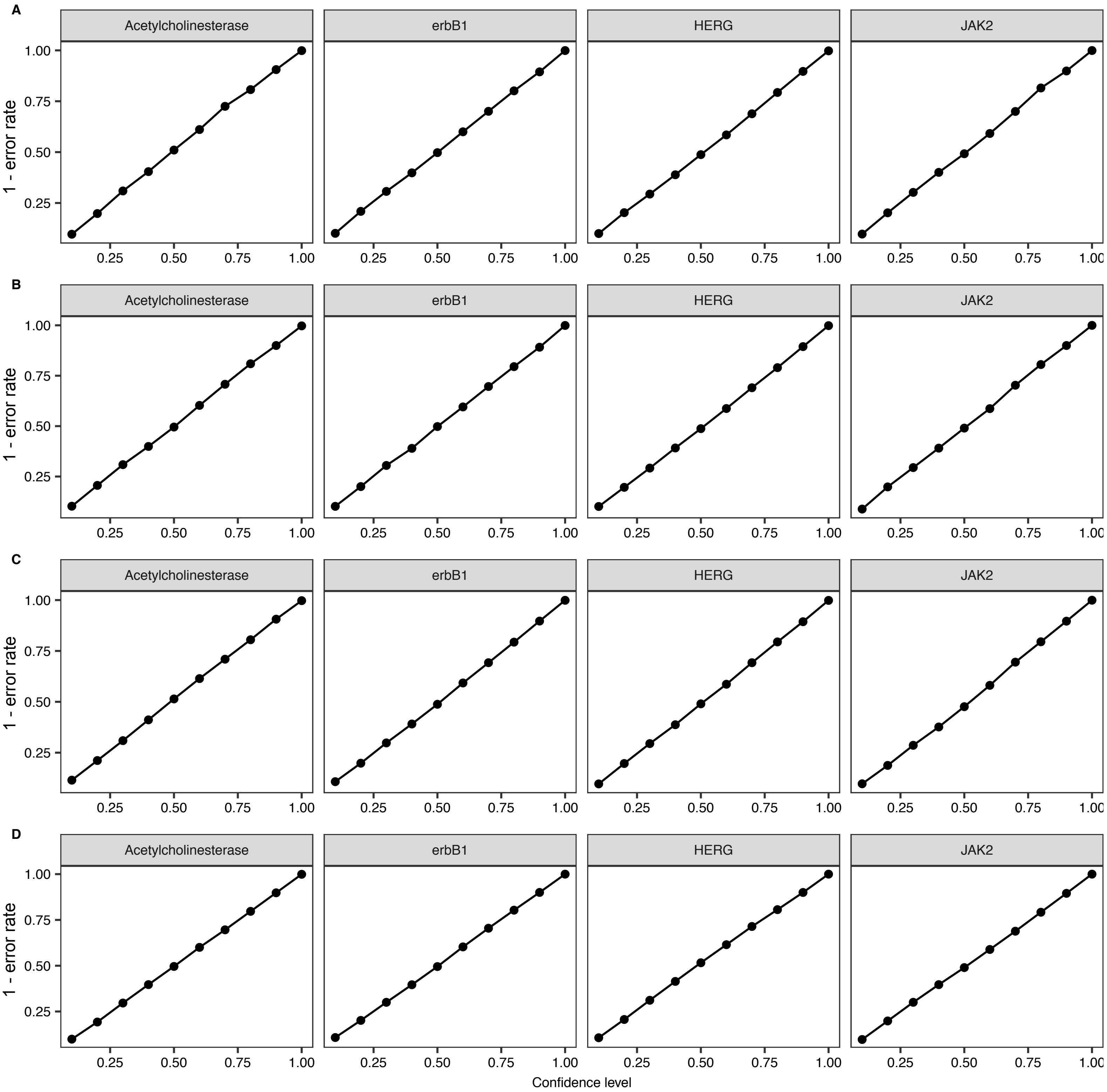

**Supplementary Figure 2**

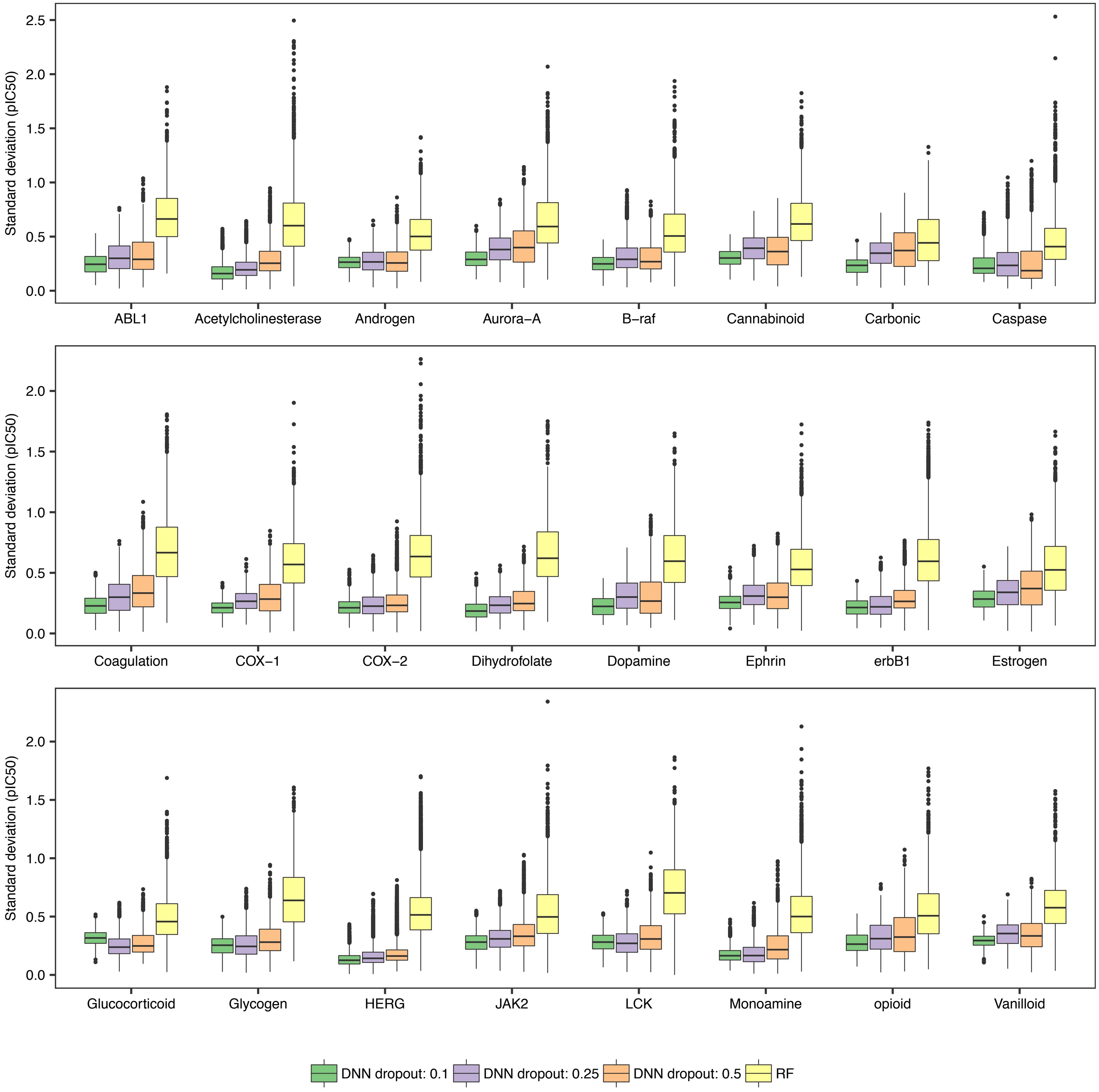